\documentclass[11pt]{article}

\usepackage[final]{acl}

\usepackage{times}
\usepackage{latexsym}
\usepackage{amsmath,amssymb}
\usepackage[T1]{fontenc}

\usepackage[utf8]{inputenc}

\usepackage{microtype}

\usepackage{inconsolata}
\usepackage{multirow}
\usepackage{graphicx}
\usepackage{booktabs}
\usepackage{amsmath}
\usepackage{graphicx}

\title{HSRM: Hidden-State Reward Models for
Test-Time Verification}

\author{
    Xianzhi Li\textsuperscript{}, 
    Xiaodan Zhu\textsuperscript{}
\\
    \textsuperscript{}Department of Electrical and Computer Engineering \& Ingenuity Labs Research Institute \\ Queen's University\\
    \{li.xianzhi, xiaodan.zhu\}@queensu.ca
}

\begin{document}
\maketitle
\begin{abstract}
Large language models can often generate plausible mathematical reasoning
traces, but reliably identifying the correct solution among multiple candidates
remains a key challenge. Existing test-time reasoning pipelines typically rely
on text-based verifiers that re-read each generated solution, making
verification an expensive component of inference. Prior work has shown,
however, that LLMs often encode correctness-related signals in their internal
representations, including awareness of when their own answers are likely to be
wrong. Building on this observation, we introduce HSRM, a lightweight
hidden-state reward model that verifies candidate solutions by directly reading
the generator's internal representations rather than re-processing its text.
HSRM extracts hidden states from a frozen generator at reasoning-step
boundaries and uses a small Transformer encoder to rank candidates. It is
trained from self-generated trajectories with outcome labels, requiring neither
human-written process supervision nor a large pretrained verifier. Across four
mathematical reasoning benchmarks, HSRM matches or outperforms a 55M-parameter
text-only energy verifier in 15 of 16 generator--dataset settings while using
only about 2M parameters, providing an efficient alternative to text-only
verification by reusing representations already computed during generation \footnote{Code is available at \url{https://github.com/JXL884/HSRM}.}.
\end{abstract}

\section{Introduction}
Large language models (LLMs) have made substantial progress on deductive reasoning, yet individual generated solutions often remain unreliable. A common way to improve reliability is to spend more computation at test time: sample multiple candidate solutions, score them, and return the most promising one. This strategy has become a standard mechanism for converting additional inference compute into higher reasoning accuracy \citep{brown2024, snell2024}. Its effectiveness, however, depends heavily on the quality and cost of the verifier used to select among candidates.

Most existing verifiers are text-based. Since the GSM8K verifier of
\citet{cobbe2021}, reward models for reasoning have typically been trained to
read a generated solution as text and assign either an outcome-level score or
process-level scores over intermediate reasoning steps
\citep{uesato2022, lightman2024, wang2024}. Recent process reward models are often large language models themselves
\citep{yang2024, zhang2025, zheng2025}. As a result, verification can become a
major part of the total inference cost: after the generator has already produced
each candidate solution, a separate model must process the same solution again in
order to judge it.

This text-only design leaves open a natural question. During generation, the
model has already computed rich internal representations that may contain
evidence about whether the solution is correct. Prior work suggests that such
information is often present in model internals: language models can encode
signals about whether their own answers are likely to be correct
\citep{kadavath2022}, and hidden representations have been used to detect
truthfulness, factuality, and correctness-related properties
\citep{azaria2023, burns2023, li2023, zou2023, marks2023, orgad2025,
zhang2025probe}. These findings suggest that a verifier may not need to infer
solution quality only from emitted text. Instead, verification could directly
read the internal states produced during decoding. This motivates our central
question: can hidden states from the generator itself support efficient
verification for mathematical reasoning?

We introduce \textbf{HSRM}, a \textbf{H}idden-\textbf{S}tate
\textbf{R}eward \textbf{M}odel for test-time verification. HSRM reads hidden
states from a frozen generator at reasoning-step boundaries and assigns a scalar
score to each candidate solution. It is implemented as a small Transformer
encoder with 2M parameters and is trained on hidden states from the
generator's own sampled trajectories. This training data is collected by the same
test-time sampling process used for best-of-\(N\) reasoning: given problems in a
target domain, we sample multiple candidate solutions, label their final answer
correctness, and train the verifier to rank correct candidates above incorrect
ones. Thus, HSRM does not rely on a pre-existing verifier corpus or
human-written process supervision. Instead, it can be adapted to a new domain
whenever candidate solutions can be sampled and outcome labels can be obtained.
Unlike text-based verifiers, HSRM does not need to re-encode generated solutions
with a separate large language model; it reuses representations that are already
produced during decoding, making verification substantially lighter. Our main contributions are:

\begin{itemize}
    \item We propose HSRM, a lightweight hidden-state reward model for
    best-of-\(N\) reasoning that scores candidate solutions using a frozen generator
    hidden states at reasoning-step boundaries and can be trained from candidate
    trajectories sampled by the generator itself.

    \item We evaluate HSRM across a variety of mathematical reasoning benchmarks with different generator scales, showing that it matches or outperforms a
    55M-parameter text-only verifier on 15 of 16 generator--dataset settings
    while using only about 2M parameters.

    \item We provide ablations over verifier capacity, layer depth, and input
    modality, showing that the hidden-state advantage comes primarily from the
    input representation rather than from increased model size.
\end{itemize}

\section{Related Work}

\paragraph{Text-based verifiers for test-time reasoning.}
Scaling test-time compute has become a standard strategy for improving LLM
reasoning.  Methods typically
sample multiple candidate solutions and select or aggregate among them
\citep{wang2023, brown2024, snell2024, wu2024}. In best-of-\(N\) reasoning,
the verifier is central: final performance depends on its ability to rank
candidate solutions. Most existing verifiers operate on generated text. Early
outcome reward models rerank complete solutions using final-answer supervision
\citep{cobbe2021}, while process reward models supervise intermediate reasoning
steps \citep{uesato2022, lightman2024, wang2024}. Recent PRMs for mathematical
reasoning are often large language models themselves
\citep{yang2024, zhang2025, zheng2025}, making verification a substantial
component of inference cost. Smaller text-based verifiers such as EORM
\citep{jiang2025} reduce this cost by training compact energy models over
chain-of-thought text, but they still require each candidate solution to be
re-encoded as text. HSRM differs by using the generator's hidden states directly,
avoiding a separate text re-encoding pass.

\paragraph{Correctness signals in model internals.}
Our work is motivated by evidence that language models encode information about
their own knowledge and correctness in internal representations. Prior work has
shown that models can exhibit signals of whether their answers are likely to be
correct \citep{kadavath2022}, and that hidden activations can support the
detection of truthfulness, factuality, and related properties
\citep{azaria2023, burns2023, li2023, zou2023, marks2023}. More recent work
further suggests that such correctness-related signals can be localized to
specific positions or internal features in long-form generations
\citep{orgad2025, zhang2025probe}. These studies indicate that hidden states may
contain information that text-only verifiers must infer indirectly from emitted
tokens. HSRM builds on this perspective by using hidden states not only for
diagnosis or probing, but as the primary input to a learned verifier.

\paragraph{Hidden-state verifiers.}
Several recent and concurrent works also explore hidden states for reasoning
verification. Some methods use probes or activation features to detect errors,
support early exiting, or perform training-free verification
\citep{zhang2025probe, liang2025clue, piotrowski2025lilave}.
SWIFT/ELHSR \citep{guo2025swift} applies a lightweight reward head to
token-level hidden states for best-of-\(N\) selection. Concurrently, ReProbe
\citep{ni2026reprobeefficienttesttimescaling} trains lightweight Transformer probes over token-level
internal features and aggregates them within each reasoning step to provide
step-level verification for test-time search. HSRM is complementary but differs
in its verification, where it extracts hidden states at step boundaries instead of token-level.

\begin{figure*}[htpb]
    \centering
    \includegraphics[width=\textwidth]{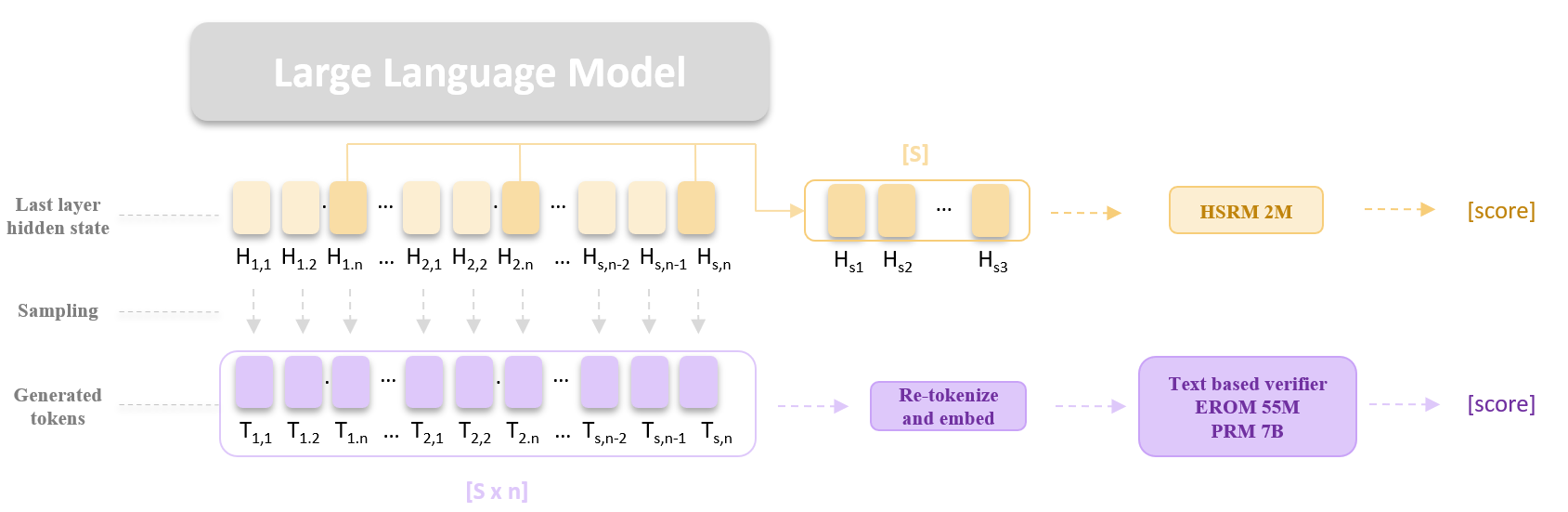}
    \caption{
\textbf{Overview of the HSRM architecture.}
During decoding, a frozen LLM produces generated tokens \(T_{s,n}\) together
with their last-layer hidden states \(H_{s,n}\). HSRM with 2M parameters extracts the hidden
states at reasoning-step boundaries to form a compact step-level sequence
\([S]\). Text-based verifiers instead take the sampled tokens, re-tokenize and
re-embed them, and run a forward pass over the full sequence \([S \times n]\)
through a much larger model. HSRM reuses representations already computed
during generation, avoiding this re-encoding step.
}
    \label{fig:headline_scaling}
\end{figure*}

\section{Our Method}
\label{sec:method}

HSRM is a lightweight hidden-state verifier for best-of-\(N\) reasoning.
Given a prompt \(x\), a frozen generator \(p_\theta\) samples \(N\) candidate
solutions \(\{y_i\}_{i=1}^N\). Instead of re-encoding each candidate as text,
HSRM reads per step hidden states produced by the generator during decoding and assigns a
scalar score to each candidate. BoN selection then returns the candidate with the highest HSRM score.

\subsection{Step-Boundary Hidden-State Extraction}
\label{sec:method:extract}

Let \(y = (y_1,\ldots,y_T)\) be a candidate solution generated for prompt \(x\).
During decoding, the generator produces hidden states
\(\{h_t^\ell\}_{t=1}^T\), where \(h_t^\ell \in \mathbb{R}^{d_{\mathrm{gen}}}\)
is the residual-stream representation at token position \(t\) and layer
\(\ell\). Rather than using all token-level hidden states, we extract
representations only at reasoning-step boundaries. We present input-modalities ablations in Section~\ref{sec:ablations:input}.

We segment each generated solution using the step stop delimiter, including
[
"\texttt{\textbackslash n\textbackslash n}",
"\texttt{.\textbackslash n}",
"\texttt{:\textbackslash n}",
"\texttt{\textbackslash n}",
"\texttt{.\textbackslash n\textbackslash n}",
"\texttt{:\textbackslash n\textbackslash n}"], which commonly separate reasoning
steps in chain-of-thought outputs. Let \(t_1 < t_2 < \cdots < t_S \leq T\)
denote the token positions immediately before these step boundaries,
including the final generated token. We collect

\begin{equation}
    H^\ell(y)
    =
    \left[
    h_{t_1}^{\ell};
    h_{t_2}^{\ell};
    \ldots;
    h_{t_S}^{\ell}
    \right]
    \in \mathbb{R}^{S \times d_{\mathrm{gen}}}.
\end{equation}

Unless otherwise stated, we use the final generator layer \(\ell=L\). In
Section~\ref{sec:ablations:layer} and \ref{sec:ablations:input}, we additionally study how performance varies across
layers. Step-boundary extraction substantially reduces the verifier input length:
instead of processing every generated token, HSRM processes only the sequence of
reasoning-step representations. Since these hidden states are already computed
during generation, HSRM does not require any additional generator forward passes.

\subsection{The HSRM Architecture}
\label{sec:method:arch}

Given a step-level hidden-state sequence
\(H \in \mathbb{R}^{S \times d_{\mathrm{gen}}}\), HSRM maps it to a scalar
candidate score through three stages: a per-step linear projection, a Transformer
encoder over reasoning steps, and a mean-pooled linear readout.

First, each generator hidden state is projected into the verifier hidden width:
\begin{equation}
    z_s^{(0)}
    =
    W_{\mathrm{in}} h_{t_s}^{\ell} + b_{\mathrm{in}}.
\end{equation}
where \(W_{\mathrm{in}} \in \mathbb{R}^{d_{\mathrm{model}} \times
d_{\mathrm{gen}}}\). The resulting sequence
\(Z^{(0)} = (z_1^{(0)}, \ldots, z_S^{(0)})\) is then passed through a small
Transformer encoder:
\begin{equation}
    Z
    =
    \mathrm{TransformerEncoder}(Z^{(0)}),
\end{equation}
where \(Z=(z_1,\ldots,z_S)\).

To obtain a candidate-level representation, HSRM mean-pools the encoded step
representations over the unpadded step positions and applies layer normalization:
\begin{equation}
    z
    =
    \mathrm{LN}
    \left(
    \frac{1}{S}
    \sum_{s=1}^{S} z_s
    \right).
\end{equation}
Finally, a single linear readout maps the pooled representation to a scalar
verifier score:
\begin{equation}
    f_\phi(x,y)
    =
    w^\top z + b .
\end{equation}

Our default HSRM uses a 2-layer Transformer encoder with hidden width
\(d_{\mathrm{model}}=256\), 4 attention heads, feed-forward width
\(4d_{\mathrm{model}}\), resulting in roughly 2M
parameters depending on the generator hidden size. We also evaluate larger
variants in Appendix~\ref{app:extabl} to separate the effect of verifier
capacity from the effect of reading hidden states.

\subsection{Training Objective}
\label{sec:method:loss}

HSRM is trained to rank correct candidates above incorrect ones within the same
problem. For each training problem, let
\(\{(y_i, c_i)\}_{i=1}^{N}\) denote the sampled candidates and their binary
correctness labels, where \(c_i \in \{0,1\}\). Let \(\mathcal{P} = \{i : c\_i = 1\}\), \(\mathcal{N} = \{j : c\_j = 0\}\) be the sets of correct and incorrect candidates, respectively. Because many candidates for the same problem can be correct, especially for
larger generators, the objective should not impose an arbitrary ordering among
correct solutions. We therefore use a tie-safe ranking loss that only requires
correct candidates to score higher than incorrect candidates. We empirically compare this objective against pointwise BCE and ListMLE in Section~\ref{sec:ablations:loss}:

\begin{equation}
\mathcal{L}_{\mathrm{rank}}
=
\frac{1}{|\mathcal{P}||\mathcal{N}|}
\sum_{i\in\mathcal{P}}
\sum_{j\in\mathcal{N}}
\log\!\left(1 + e^{-(s_i-s_j)}\right).
\label{eq:pairwise-rank}
\end{equation}

where \(s_i = f_\phi(x,y_i)\). This objective encourages every correct candidate
to outrank incorrect candidates, while treating all correct candidates as
ties with respect to one another. Problems with
\(|\mathcal{P}| = 0\) or \(|\mathcal{N}| = 0\) provide no within-problem ranking
signal and are omitted from the training loss.

In practice, we train HSRM on cached hidden-state tensors. The generator is run
once to produce candidate solutions and their step-boundary hidden states. HSRM
is then optimized on these cached representations without further generator
calls.

\subsection{Inference}
\label{sec:method:inference}

At inference time, the generator samples \(N\) candidate solutions for each
problem and caches the same step-boundary hidden states used during training.
HSRM scores the \(N\) candidates in a batched verifier forward pass, and BoN
selection returns the highest-scoring candidate:
\begin{equation}
    \hat{y}
    =
    y_{\arg\max_i f_\phi(x,y_i)} .
\end{equation}
Because HSRM operates over step-level hidden states that are already produced
during generation, it avoids the additional generator or verifier-side text
re-encoding required by text-based verifiers. Its added inference cost is
therefore limited to a small forward pass over the extracted step representations. We provide detailed efficiency analysis in Section \ref{sec:ablations:eff}.




\section{Experimental Setup}
\label{sec:setup}

\subsection{Generators and Datasets}
\label{sec:setup:data}

We evaluate HSRM using Qwen3 generators \citep{qwen3} at four scales:
1.7B, 4B, 8B, and 14B. Our main experiments use all generators in non-thinking mode and keep them
frozen throughout training and evaluation; only HSRM and the learned verifier
baselines are trained. We additionally evaluate Qwen3 in thinking mode in
Section~\ref{sec:ablations:thinking} to study how explicit deliberation affects
hidden-state extraction.

We consider four mathematical reasoning benchmarks: GSM8K \citep{cobbe2021},
MATH-500 \citep{hendrycks2021measuringmathematicalproblemsolving}, AIME \citep{aopsAIME}, and OlympiadBench
\citep{he2024olympiadbench}. For each dataset, we construct a verifier training
pool by sampling 64 candidate solutions per training problem. At evaluation
time, we report best-of-8 selection on a disjoint evaluation split. Exact split indices are provided in Appendix~\ref{app:splits}.

\subsection{Candidate Labeling}
\label{sec:setup:labels}

Each candidate solution is assigned a binary correctness label. We first extract
the final answer using dataset-specific rules and compare it against the ground
truth after normalizing common formatting differences and equivalent
symbolic expressions. Ambiguous cases are resolved using an LLM judge that is
given the ground-truth answer as context. Unless otherwise stated, all reported
results use these post-processed correctness labels. Detailed relabeling
statistics are provided in Appendix~\ref{app:relabel}.

\subsection{Baselines}
\label{sec:setup:baselines}

We compare HSRM against three classes of baselines. First, we compare against text-based reward models. Our main text-only baseline
is EORM \citep{jiang2025}, an energy-based verifier that reads the full
chain-of-thought text and scores candidates with a Transformer encoder trained
using a Bradley--Terry ranking objective. We use EORM 55M as the primary text-only comparison.

Second, we compare against Qwen2.5-Math-PRM-7B \citep{zhang2025}, a 7B
off-the-shelf process reward model trained with MATH-style process supervision.
We treat this model as a strong external PRM baseline rather than a
capacity-matched comparison.

Lastly, we evaluate non-learned generator-internal scoring heuristics, including single-pass
generation and oracle
pass@\(N\). We also report simple baselines like cumulative
log-probability, mean log-probability, negative mean entropy, negative
varentropy, response length, number of reasoning steps, and longest response in Table \ref{tab:app_cheap}. These baselines test whether simple confidence, uncertainty, or length-based
signals are sufficient for best-of-\(N\) selection.

\subsection{Training and Evaluation}
\label{sec:setup:training_eval}

Unless otherwise stated, HSRM uses a 2-layer Transformer encoder with hidden
width 256, 4 attention heads, dropout 0.1, and final-layer step-boundary hidden
states. More training details can be found in Appendix~\ref{app:arch}. All learned verifiers are trained with five random seeds, using a
problem-level validation split for early stopping. HSRM is trained on cached
hidden-state tensors: after the candidate pools and step-boundary hidden states
are generated once, verifier training requires no additional generator calls.

Our primary metric is verifier-best accuracy at \(N=8\). For each problem, the
verifier selects the highest-scoring candidate among eight sampled solutions, and
the prediction is counted as correct if the selected candidate has the correct
final answer. We also report within-problem AUROC, computed by comparing
verifier scores against binary correctness labels within each candidate pool and
then averaging over problems for which AUROC is defined. Oracle pass@\(N\) is
reported as the upper bound imposed by the sampled candidate pool.

\begin{figure*}[htpb!]
    \centering
    \includegraphics[width=\textwidth]{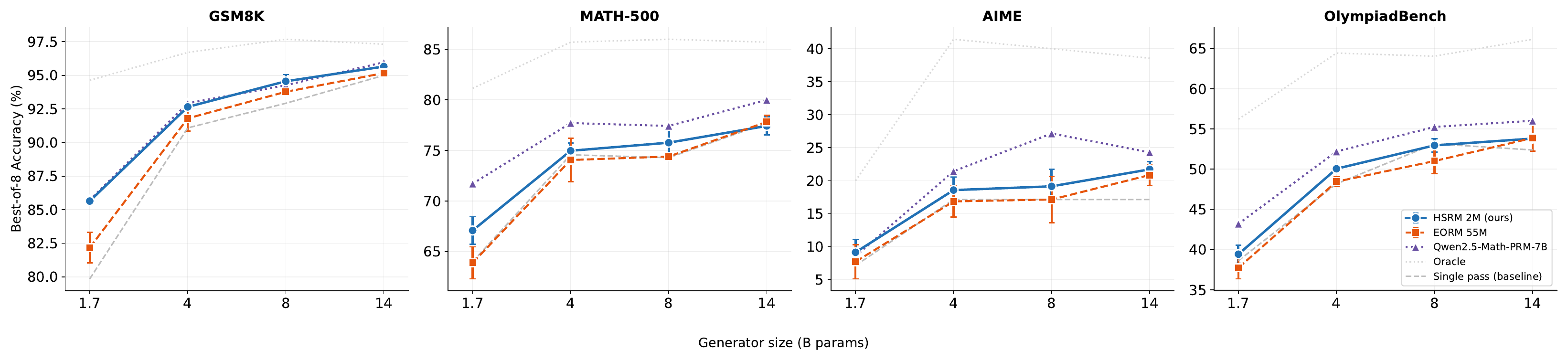}
    \caption{
    Best-of-8 verifier accuracy as generator size increases. HSRM improves
    over the 55M-parameter text-only EORM baseline across most generator--dataset
    settings while using only about 2M parameters. The oracle curve shows the
    pass@8 upper bound, and the single-pass curve shows the accuracy without
    verifier-based selection. Results are reported as mean \(\pm\) standard
    deviation over five seeds.
    }
    \label{fig:headline_scaling}
\end{figure*}

In-distribution results are reported as mean \(\pm\) standard deviation over five seeds. For zero-shot transfer, we train the verifier on a source dataset
and evaluate it directly on a target dataset, without any target-domain verifier
training.

\section{Results}
\label{sec:results}

\subsection{Main Results}
\label{sec:results:main}

We first evaluate whether HSRM can serve as an effective verifier for
best-of-\(N\) selection across generator scales and benchmark difficulty.
Figure~\ref{fig:headline_scaling} reports best-of-8 accuracy for Qwen3
generators from 1.7B to 14B parameters on GSM8K, MATH-500, AIME, and
OlympiadBench. We compare HSRM with EORM, a 55M-parameter text-only energy
verifier, as well as single-pass generation, oracle pass@8, and
Qwen2.5-Math-PRM-7B.

On GSM8K, HSRM outperforms EORM at every generator scale and nearly matches
Qwen2.5-Math-PRM-7B, despite using only about 2M parameters instead of a
billion-parameter text-based verifier. As the generator scale increases, HSRM also
approaches the oracle curve, indicating that GSM8K is close to saturated in the
best-of-8 setting: correct candidates are often present, and HSRM can identify
them nearly as effectively as a much larger domain-trained PRM. This suggests that
generator hidden states contain useful information for distinguishing correct
from incorrect candidates, especially before the generator saturates the task.

The benefits extend to harder benchmarks. On MATH-500, AIME, and OlympiadBench,
HSRM generally remains above EORM while improving with generator scale.
Qwen2.5-Math-PRM-7B retains an advantage on these more challenging mathematical
benchmarks, especially AIME, which is expected given its much larger scale and
domain-specific training. Nevertheless, HSRM closes a substantial portion of the
gap while avoiding the cost of re-encoding candidate solutions as text.

Overall, Figure~\ref{fig:headline_scaling} supports our main claim: compact
hidden-state verification can outperform a substantially larger text-only
verifier across generator scales and benchmark difficulty, and can match a large
domain-trained PRM on GSM8K. HSRM is most attractive when verifier efficiency is
important, while large domain-matched PRMs remain stronger on the hardest
benchmarks.

\section{Ablations} \label{sec:ablations}

\begin{figure*}[t]
    \centering
    \includegraphics[width=\textwidth]{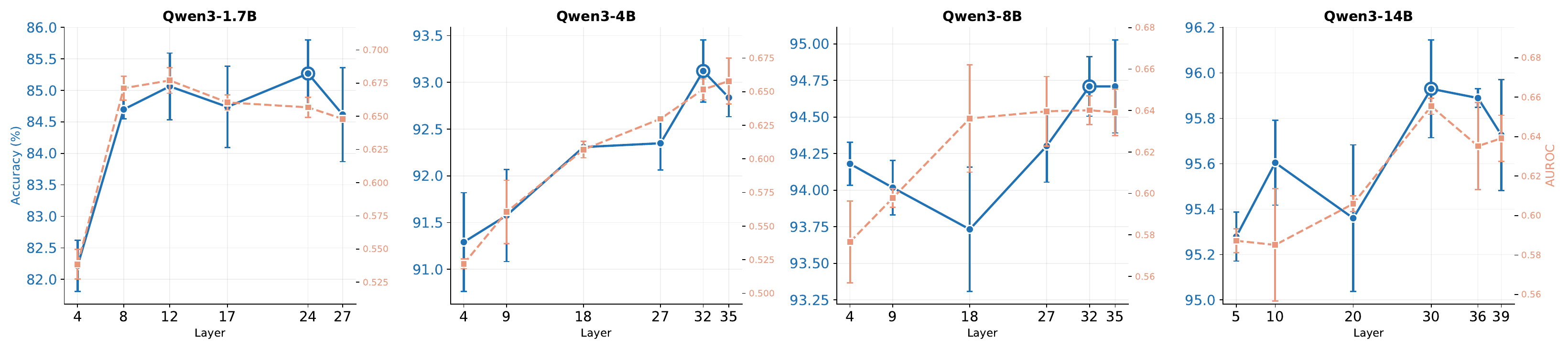}
    \caption{
    Layer ablation for HSRM on GSM8K. We vary the generator layer used for
    step-boundary hidden-state extraction and report both best-of-8 accuracy
    and within-problem AUROC. Correctness information is present across many
    layers, but the strongest performance generally occurs in the upper portion
    of the generator. Results are reported as mean \(\pm\) standard deviation
    over three seeds.
    }
    \label{fig:layer_ablation}
\end{figure*}

\begin{table*}[t]
\centering
\small
\setlength{\tabcolsep}{4.5pt}
\begin{tabular}{lccccccc}
\toprule
Input modality & Verifier input & Generator layers & Input dim. & \(d_{\mathrm{model}}\) & Params & Bo8 Acc. & Within-AUROC \\
\midrule
Hidden state & Last hidden state  & Last layer & 2048 & 256 & 2.12M
& \(86.11 \pm 0.35\) & \(0.691 \pm 0.005\) \\
Hidden state & Step hidden state & Last layer & 2048 & 256 & 2.12M
& \(86.32 \pm 1.08\) & \(0.714 \pm 0.018\) \\
Hidden state & Last hidden state & Top-4 layers & 8192 & 256 & 3.69M
& \(86.28 \pm 0.88\) & \(0.715 \pm 0.015\) \\
Hidden state & Step hidden state & Top-4 layers & 8192 & 256 & 3.69M
& \(\mathbf{86.86 \pm 0.74}\) & \(\mathbf{0.724 \pm 0.009}\) \\
Text only & Raw solution text & -- & -- & -- & 2.56M
& \(82.23 \pm 0.29\) & \(0.519 \pm 0.009\) \\
Hybrid & Text + hidden states & Top-4 layers & 8192 & 256 & 4.77M
& \(86.23 \pm 0.60\) & \(0.707 \pm 0.021\) \\
\bottomrule
\end{tabular}
\caption{
Input-modality ablation on GSM8K using Qwen3-1.7B in non-thinking mode.
Hidden-state verifiers read step-boundary representations, extracted from the
token immediately before each step stop delimiter. Last-layer models use a 2048-dimensional
input, while top-4-layer models concatenate layers \(-1,-2,-3,-4\), giving an
8192-dimensional input. The text-only baseline encodes the raw solution text with
a from-scratch encoder, and the hybrid model fuses text and hidden-state inputs.
}
\label{tab:input_modality_ablation}
\end{table*}

\subsection{Generator Layer}
\label{sec:ablations:layer}

HSRM uses hidden states from the frozen generator as its input, so an important
design question is which generator layer should be used for verification. Prior
work on contextual representations suggests that useful information is often
distributed across layers rather than concentrated only in the final layer:
different layers encode different linguistic and semantic abstractions, and
intermediate or near-final layers can sometimes provide more transferable
features than the final representation \citep{peters2018deep, liu2019linguistic, skean2025uncovering}. This is especially
relevant for hidden-state verification, since the final layer is also the layer
most directly shaped for next-token prediction, whereas earlier upper layers may
retain richer information about the reasoning trajectory.

We therefore evaluate HSRM while varying the generator layer used for
step-boundary hidden-state extraction. Figure~\ref{fig:layer_ablation} reports
both best-of-8 accuracy and within-problem AUROC on GSM8K for Qwen3 models from
1.7B to 14B parameters. The results show that correctness information is present
across a broad range of layers, with the strongest performance generally
appearing in the upper portion of the generator. The final layer is competitive, but it is not always the
unique optimum. In several settings, nearby upper layers match or exceed the
final layer, suggesting that verification-relevant information is distributed
across multiple high-level representations.

We use the final layer as the default in the main experiments because it is
simple, consistently competitive, and avoids tuning a layer choice separately
for each generator--dataset pair. At the same time, the layer ablation motivates
multi-layer hidden-state inputs. In Section~\ref{sec:ablations:input}, we therefore
also test a top-4-layer variant that concatenates the final four generator
layers; this setting gives the strongest overall ranking performance, supporting
the view that hidden-state verification benefits from combining information
across multiple upper layers rather than relying exclusively on a single final
representation.

\subsection{Input Modality}
\label{sec:ablations:input}

Table~\ref{tab:input_modality_ablation} studies which input signal is most
useful for HSRM. The text-only verifier performs substantially worse than all
hidden-state variants, reaching only \(82.23\%\) best-of-8 accuracy and
\(0.519\) within-problem AUROC. In contrast, even the simplest hidden-state
verifier, which reads only the final step representation from the last generator
layer, achieves \(86.11\%\) accuracy and \(0.691\) AUROC. This gap suggests that
the improvement does not come merely from the verifier architecture, but from
the hidden-state representation itself.

Using the full sequence of step-boundary representations further improves
ranking quality, and concatenating the top four generator layers gives the best
overall result, reaching \(86.86\%\) accuracy and \(0.724\) AUROC. This supports
the conclusion from the layer ablation: correctness information is not confined
to the final hidden state, but is distributed both across the reasoning
trajectory and across multiple upper layers of the generator. The AUROC gains
are especially important because best-of-\(N\) selection depends on ranking
correct and incorrect candidates within the same problem.

Interestingly, the hybrid text-plus-hidden model does not improve over the
hidden-only model despite using more parameters. This indicates that the
generator's internal representations already provide a strong verification
signal, and adding raw text features can introduce extra complexity without
improving selection performance.

\subsection{Efficiency Analysis} \label{sec:ablations:eff}

\begin{figure}[t]
    \centering
    \includegraphics[width=\columnwidth]{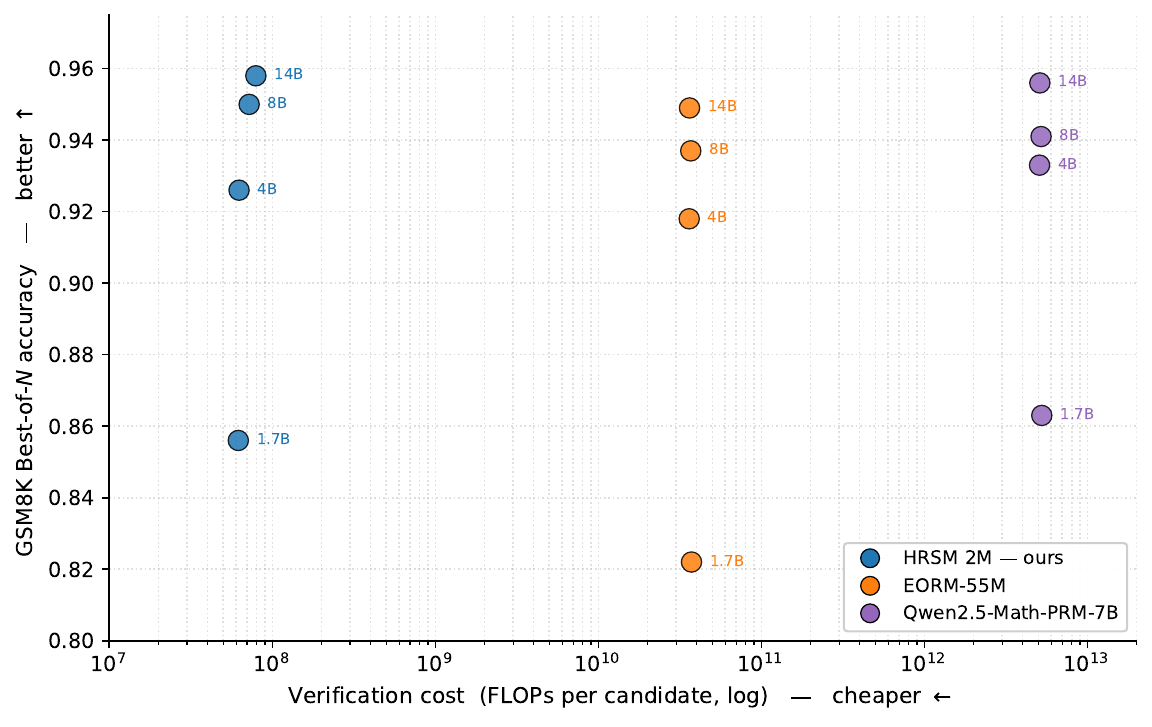}
    \caption{
    Verification efficiency frontier on GSM8K.
    Best-of-\(N\) accuracy versus estimated verification FLOPs per candidate
    (log scale) for HSRM, EORM-55M, and Qwen2.5-Math-PRM-7B across four
    generator scales. HSRM lies on the
    upper-left frontier, matching the 7B PRM at the larger
    generator scales at roughly five orders of magnitude lower verification
    cost.
    }
    \label{fig:efficiency-frontier}
\end{figure}

In best-of-\(N\) reasoning, verifier cost scales directly with the number of
sampled candidates, making per-candidate efficiency central to test-time
compute. Figure~\ref{fig:efficiency-frontier} compares HSRM with text-based verifiers, plotting best-of-\(N\) accuracy against estimated
verification FLOPs per candidate. HSRM lies on the upper-left efficiency frontier
across generator scales. This gain comes with a substantial cost advantage: HSRM uses about \(3500\times\)
fewer parameters than the 7B PRM and roughly five orders of magnitude fewer
verification FLOPs per candidate, since it reads cached hidden states rather
than re-encoding each solution with a large text model. This difference is
especially important in best-of-\(N\) settings, where the same verifier must be
applied repeatedly to many sampled candidates. By moving verification from
full-sequence text re-encoding to a compact step-level hidden-state reader, HSRM
makes additional test-time samples more affordable. These results show that HSRM
improves not only parameter efficiency, but also the accuracy--cost frontier for
test-time verification.

\begin{figure}[t]
    \centering
    \includegraphics[width=0.5\textwidth]{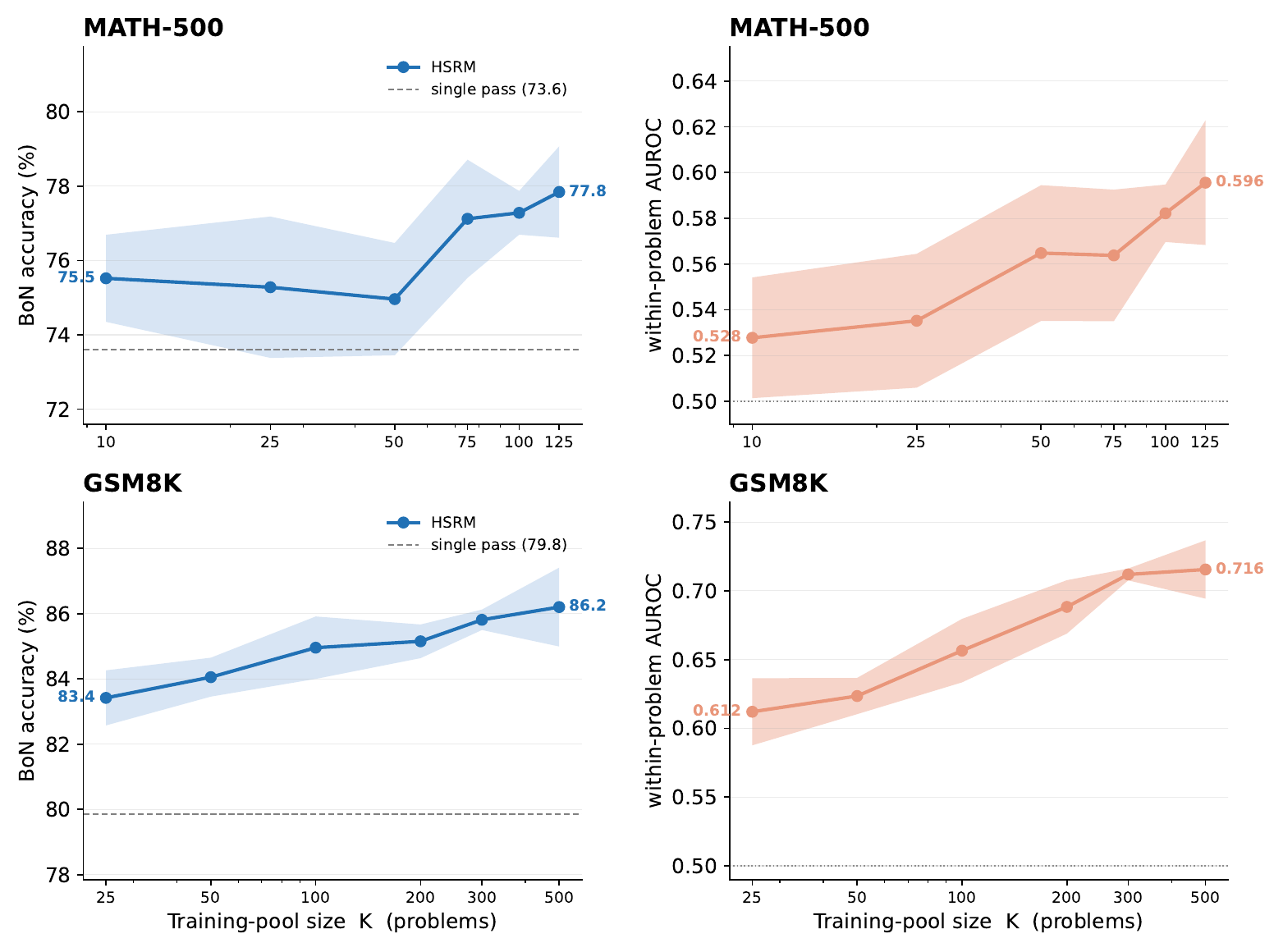}
    \caption{
    Scaling with verifier training-pool size \(K\) for HSRM on Qwen3-1.7B.
    We vary the number of training problems used to train the verifier and
    evaluate on a disjoint test split for both MATH-500 and GSM8K. Left:
    best-of-8 accuracy. Right: within-problem AUROC. HSRM improves with more
    training data on both datasets, with especially consistent gains on GSM8K.
    Shaded regions show mean \(\pm\) standard deviation over five seeds.
    }
    \label{fig:poolsize_ablation}
\end{figure}

\subsection{Training Data}
\label{sec:ablations:poolsize}

We next study how HSRM scales with the size of the verifier training pool.
Figure~\ref{fig:poolsize_ablation} varies the number of training problems \(K\)
for a Qwen3-1.7B generator, with evaluation problems kept disjoint from the
training pool. HSRM generally improves as \(K\) increases. On GSM8K, increasing
\(K\) from 25 to 500 raises best-of-8 accuracy from 83.4 to 86.2 and
within-problem AUROC from 0.612 to 0.716. MATH-500 shows the same overall
direction, with accuracy increasing from 75.5 to 77.8 and AUROC from 0.528 to
0.596 as \(K\) grows from 10 to 125, although the curve is noisier due to the
harder and more heterogeneous problem distribution. Since best-of-\(N\)
reasoning relies on a verifier to select among sampled candidates, these gains
indicate that additional verifier data mainly strengthens the ranking signal.
At the same time, the curve shows that
small training pools already provide useful gains over single-pass generation,
while larger pools offer better ranking quality at the cost of more data
collection and labeling. Thus, the choice of \(K\) is application-dependent:
one can use a small pool for a cheap verifier adaptation, or increase \(K\) when
higher ranking accuracy is worth the additional supervision cost.

\subsection{Zero-Shot Transfer}
\label{sec:ablations:transfer}

\begin{figure}[ht]
    \centering
    \includegraphics[width=0.5\textwidth]{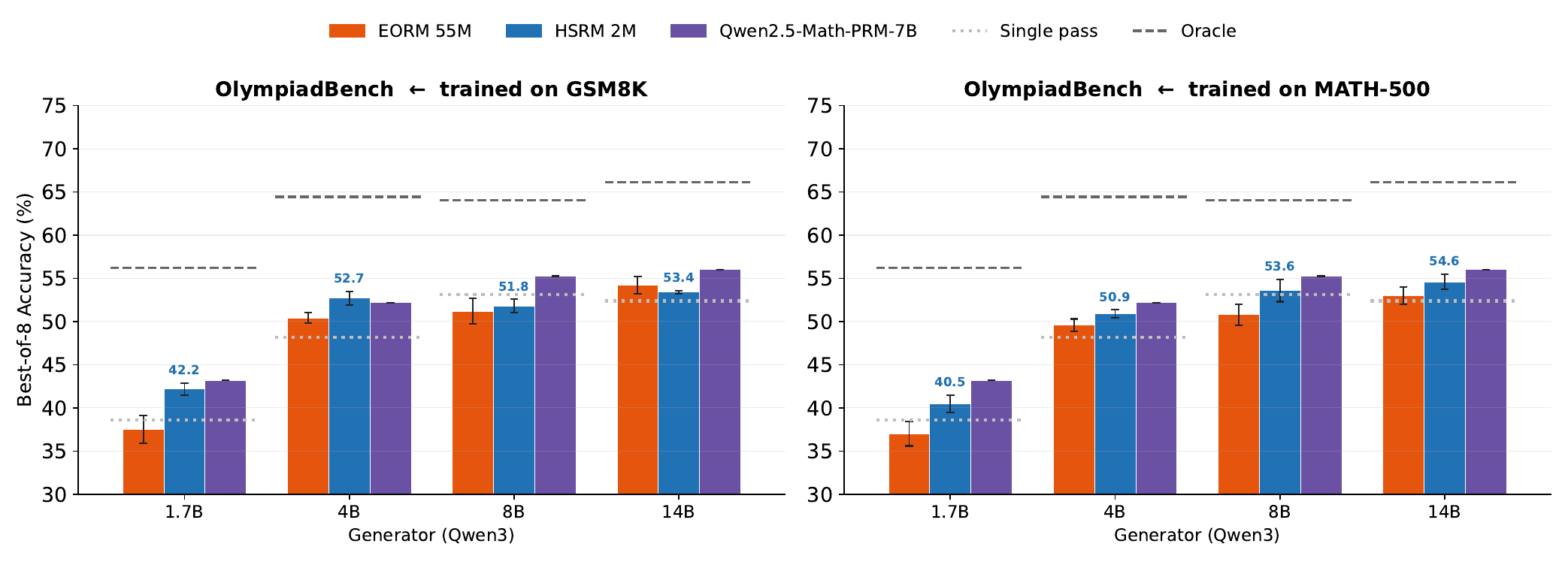}
    \caption{
    Zero-shot transfer to OlympiadBench. Verifiers are trained only on GSM8K
    (left) or MATH-500 (right), and evaluated directly on OlympiadBench without
    target-domain tuning. HSRM consistently outperforms the text-only EORM 55M
    baseline and remains close to Qwen2.5-Math-PRM-7B across generator scales. Dotted
    lines show single-pass accuracy and dashed lines show oracle pass@8.
    Results are mean \(\pm\) standard deviation over five seeds.
    }
    \label{fig:transfer_olympiadbench}
\end{figure}

We test whether HSRM learns source-specific dataset patterns or more
general verification signals. We train verifiers only on GSM8K or MATH-500 and
evaluate them directly on OlympiadBench, without any target-domain training.
Figure~\ref{fig:transfer_olympiadbench} reports best-of-8 accuracy across
Qwen3 generator scales.

HSRM transfers well under this distribution shift. When trained on GSM8K, HSRM
outperforms EORM 55M at every generator scale and remains close to Qwen2.5-Math-PRM-7B,
despite being much smaller. The same pattern holds when training on MATH-500:
HSRM again improves over the text-only verifier across all scales and tracks the
large PRM baseline closely. This suggests that the hidden-state ranking signal
learned by HSRM is not tied only to the source dataset, but captures reusable
features of reasoning quality. Therefore, transfer studies provide evidence that generator hidden states
encode verification-relevant signals that can transfer across mathematical
reasoning domains, even without target-domain verifier training.

The remaining gap is primarily to the oracle rather than between learned
verifiers. On OlympiadBench, oracle pass@8 is much higher than the selected
accuracy for all methods, meaning that many candidate pools contain a correct
solution that the verifier does not select. At the same time, the gap between
HSRM and Qwen2.5-Math-PRM-7B is relatively small compared with the oracle headroom.
Thus, zero-shot transfer performance appears limited less by the choice between
small hidden-state and large text-based verifiers, and more by the difficulty of
generating and reliably selecting correct candidates on OlympiadBench.

\subsection{Thinking-Mode Generators}
\label{sec:ablations:thinking}

Our main experiments use Qwen3 in non-thinking mode, where the generated
reasoning trace and final answer form a single sequence. We additionally study
HSRM with Qwen3's thinking mode, which introduces an explicit internal
deliberation phase before the final answer. This setting raises a new extraction
question: should the verifier read hidden states throughout the full reasoning
trace, or only from the post-deliberation answer segment?

We compare two variants on MATH-500 using Qwen3-1.7B and Qwen3-4B.
\textit{Full-trace} extraction applies the same step-boundary reader across the
entire generated sequence, including the thinking trace. \textit{Answer-only}
extraction instead retains only representations after the \texttt{</think>}
boundary. For reference, we also report the corresponding non-thinking
generator with answer-only extraction.

\begin{table*}[t]
\centering
\small
\setlength{\tabcolsep}{4.5pt}
\begin{tabular}{llccc}
\toprule
Generator & Mode & Input & Acc & Within-AUROC \\
\midrule
\multirow{3}{*}{Qwen3-1.7B}
& Non-thinking & answer-only & $67.14 \pm 0.81$ & $0.631 \pm 0.025$ \\
& Thinking & full-trace      & $82.00 \pm 0.82$ & $0.669 \pm 0.027$ \\
& Thinking & answer-only     & $\mathbf{83.67 \pm 0.94}$ & $\mathbf{0.736 \pm 0.034}$ \\
\midrule
\multirow{3}{*}{Qwen3-4B}
& Non-thinking & answer-only & $77.62 \pm 0.49$ & $0.657 \pm 0.010$ \\
& Thinking & full-trace      & $86.33 \pm 1.70$ & $0.672 \pm 0.044$ \\
& Thinking & answer-only     & $\mathbf{87.00 \pm 0.00}$ & $\mathbf{0.884 \pm 0.023}$ \\
\bottomrule
\end{tabular}
\caption{
Thinking-mode extraction ablation on MATH-500. For thinking-mode generators,
\textit{full-trace} uses step-boundary hidden states from both the deliberation
trace and final answer, whereas \textit{answer-only} retains representations
after the \texttt{</think>} boundary. Restricting extraction to the
post-deliberation answer improves within-problem ranking quality at both model
scales.
}
\label{tab:thinking_mode}
\end{table*}

Table~\ref{tab:thinking_mode} shows two complementary effects. First, enabling
thinking substantially improves best-of-8 performance over the corresponding
non-thinking generators. More importantly for HSRM, however, using the entire
thinking trace is not the strongest way to read these models. Restricting HSRM
to post-deliberation representations improves within-problem AUROC from
$0.669$ to $0.736$ for Qwen3-1.7B and from $0.672$ to $0.884$ for Qwen3-4B.

This result suggests that verification-relevant information is not equally
stable throughout an explicit deliberation trace. During thinking, the model
may temporarily represent an incorrect hypothesis, uncertainty, or a detected
error before subsequently revising its reasoning. A hidden state at such a
step can therefore reflect the local status of an intermediate trajectory
rather than the correctness of the final answer. Representations after the
\texttt{</think>} boundary are produced after this self-correction process and
are consequently more predictive of final-answer correctness.

\subsection{Cross-Family Generalization}
\label{sec:results:llama}

The main experiments above use Qwen3 generators. To test whether HSRM relies
on properties specific to the Qwen model family, we additionally evaluate it
on candidate solutions generated by three Llama models:
Llama-3.2-1B, Llama-3.2-3B, and Llama-3.1-8B \citep{grattafiori2024llama3herdmodels}. We consider both GSM8K and
MATH-500 and train an HSRM and matched EORM verifier on candidate pools
generated by each model.

\begin{table}[t]
\centering
\small
\setlength{\tabcolsep}{3.2pt}
\begin{tabular}{llccc}
\toprule
Dataset & Generator & HSRM & EORM & Oracle \\
\midrule
\multirow{3}{*}{GSM8K}
& Llama-3.2-1B & \textbf{48.4} & 40.3  & 72.2 \\
& Llama-3.2-3B & \textbf{81.1} & 75.1  & 93.0 \\
& Llama-3.1-8B & \textbf{88.3} & 85.2  & 96.0 \\
\midrule
\multirow{3}{*}{MATH-500}
& Llama-3.2-1B & \textbf{30.1} & 26.5 & 61.4 \\
& Llama-3.2-3B & \textbf{47.9} & 43.4 & 76.3 \\
& Llama-3.1-8B & \textbf{56.8} & 52.3 & 87.1 \\
\bottomrule
\end{tabular}
\caption{
Best-of-8 accuracy (\%) with Llama-family generators.
HSRM consistently outperforms the matched text-based EORM verifier across both datasets and all three generator scales.
Oracle denotes pass@8 accuracy of the sampled candidate pool.
}
\label{tab:llama_results}
\end{table}

Table~\ref{tab:llama_results} shows that the advantage of hidden-state
verification extends beyond the Qwen family. HSRM outperforms the matched
text-based EORM verifier in all six Llama generator--dataset settings. On
GSM8K, the improvement ranges from $3.1$ to $8.1$ percentage points, while on
MATH-500 it ranges from $3.6$ to $4.5$ points.

These cross-family results provide evidence that the verification signal
exploited by HSRM is not tied to a particular generator architecture or
pretraining recipe. Despite differences between Qwen3 and Llama models, a
small verifier operating directly on their internal representations
consistently provides stronger candidate ranking than re-encoding the
generated reasoning text with the matched EORM baseline.

\section{Conclusion}
\label{sec:conclusion}

We presented HSRM, a lightweight hidden-state verifier for test-time
mathematical reasoning. Instead of re-encoding candidate solutions with a
separate text-based reward model, HSRM ranks candidates using hidden states
already produced by a frozen generator during decoding. Across four benchmarks
and both Qwen and Llama generators, HSRM consistently matches or outperforms a substantially
larger text-only energy verifier while using only about 2M parameters. Our ablation studies further show that verification-relevant information is
distributed across upper generator layers, and that HSRM offers a favorable
trade-off between training data and test-time efficiency. Together, HSRM offers a low-cost complement to existing
verifier-based test-time reasoning pipelines.

\section*{Limitations}

Our study focuses on mathematical reasoning. This controlled setting allows us
to isolate the verification signal present in the generator's hidden states,
but future work should evaluate whether the same approach extends to other
domains and a broader range of model architectures. While our experiments
include both Qwen and Llama generators and evaluate Qwen3 in thinking mode,
the thinking-mode study is limited to two model scales on MATH-500, and broader
evaluation of explicit deliberation remains an important direction.



\bibliography{custom}

\clearpage
\appendix
\label{sec:appendix}

\appendix

\section{Dataset Details and Split Indices}
\label{app:splits}

Table~\ref{tab:app_splits} reports the problem-index ranges and candidate counts
used throughout the paper. All training and evaluation splits are disjoint at the
problem level. For each training problem, we sample $N{=}64$ candidate solutions;
for each evaluation problem, we use $N{=}8$ candidates for best-of-$N$ selection.
All Qwen3 generators are run in non-thinking mode with \texttt{float16},
temperature $0.7$, and top-$p$ $0.9$. Across all datasets and splits, the
post-processed candidate corpus contains approximately $255$K sampled solutions
(Appendix~\ref{app:relabel}).

\section{Candidate Labeling and Relabeling Statistics}
\label{app:relabel}

Each sampled candidate solution is assigned a binary final-answer correctness
label before verifier training and evaluation. We use a two-stage labeling
procedure designed to preserve high-precision symbolic matches while recovering
correct answers that are missed by simple extraction rules.

In the first stage, we apply a deterministic regex/symbolic labeler. The labeler
extracts a candidate's final answer using a cascade of answer patterns, including
boxed answers, phrases such as ``the answer is'', more generic answer statements,
and, as a fallback, the last numeric expression in the solution. The extracted
answer is then compared with the ground-truth answer after normalizing common
formatting differences. Numeric answers are compared with a small tolerance, and
symbolic answers are checked with mathematical equivalence tools when possible;
otherwise, normalized string matching is used. If this stage extracts an answer
and verifies it as equivalent to the ground truth, the candidate is marked
correct and the label is locked in.

The second stage handles candidates not resolved as correct by the deterministic
labeler. This includes cases where answer extraction fails, where the extracted
span is incomplete, or where the answer contains formats that are difficult to
verify with simple rules, such as fractions, square roots, embedded units,
symbolic expressions, or multiple plausible final-answer spans. These unresolved
or provisionally incorrect candidates are batched and passed to an LLM judge,
which receives the problem, the candidate solution, and the ground-truth answer.
The judge determines whether the candidate's final answer is mathematically
equivalent to the ground truth, allowing us to recover correct solutions that
would otherwise be counted as false negatives by the regex/symbolic labeler.

This relabeling stage has a non-negligible effect on the final supervision. Over
approximately \(255\)K sampled candidate solutions, the LLM-judge stage changed
\(7{,}788\) labels in total. It corrected \(7{,}650\) candidates from incorrect
to correct and \(138\) candidates from correct to incorrect, yielding a net
increase of \(7{,}512\) correct labels. Thus, about \(3.1\%\) of candidate labels
were changed by the judge, with most changes recovering correct answers that the
deterministic labeler failed to recognize. This asymmetry is expected: strict
regex and symbolic checks are high precision but can miss semantically correct
answers written in non-canonical forms.

All reported training and evaluation results use the post-relabel correctness
labels. This ensures that HSRM is trained and evaluated against a more complete
notion of final-answer correctness, rather than against artifacts of a particular
answer-extraction rule.

\section{Architecture and Hyperparameter Details}
\label{app:arch}

\paragraph{Parameter counts.}
Table~\ref{tab:app_params} gives the parameter count of the default HSRM verifier
for each generator scale. The only component whose size depends on the generator
is the input projection $W_{\mathrm{in}} \in \mathbb{R}^{d_{\mathrm{model}} \times
d_{\mathrm{gen}}}$. The step-level Transformer encoder and final score head are
shared across generator scales.

\begin{table}[ht!]
\caption{\textbf{HSRM parameter counts by generator.} The default verifier uses
$d_{\mathrm{model}}{=}256$, two Transformer layers, four attention heads, and
dropout $0.1$.}
\label{tab:app_params}
\centering
\small
\begin{tabular}{lcc}
\toprule
Generator & $d_{\mathrm{gen}}$ & Params \\
\midrule
Qwen3-1.7B & 2048 & 2.12M \\
Qwen3-4B   & 2560 & 2.25M \\
Qwen3-8B   & 4096 & 2.65M \\
Qwen3-14B  & 5120 & $\sim$3.4M \\
\bottomrule
\end{tabular}
\end{table}

\begin{table*}[ht]
\caption{\textbf{Dataset splits and candidate counts.} ``Source'' denotes the
underlying benchmark split. MATH-500 uses the first 150 problems for verifier
training and the remaining 350 problems as the strict holdout used for headline
results.}
\label{tab:app_splits}
\centering
\small
\begin{tabular}{llrlrl}
\toprule
Dataset & Source split & Train idx. & ($n$, $N$) & Eval idx. & ($n$, $N$) \\
\midrule
GSM8K          & test ($1319$) & $[0,499]$   & ($500$, $64$) & $[500,1318]$ & ($819$, $8$) \\
MATH-500       & test ($500$)  & $[0,149]$   & ($150$, $64$) & $[150,499]$  & ($350$, $8$) \\
AIME           & aimo-val ($90$) & $[0,19]$  & ($20$, $64$)  & $[20,89]$    & ($70$, $8$) \\
OlympiadBench  & \texttt{OE\_TO\_maths\_en\_COMP} ($673$) & $[0,149]$ & ($150$, $64$) & $[150,672]$ & ($523$, $8$) \\
\bottomrule
\end{tabular}
\end{table*}

\paragraph{Training hyperparameters.}
Table~\ref{tab:app_hparams} summarizes the verifier training recipe. Unless
otherwise stated, we use the same hyperparameters across datasets and generator
scales.

\begin{table*}[t]
\caption{\textbf{HSRM training hyperparameters.}}
\label{tab:app_hparams}
\centering
\small
\begin{tabular}{ll}
\toprule
Hyperparameter & Value \\
\midrule
Optimizer & AdamW \\
Learning rate & $1\times10^{-4}$ \\
Problem batch size & $8$ \\
Gradient steps & $1000$ \\
Validation split & $0.20$ problem-level validation split, seed $42$ \\
Dropout & $0.1$ \\
Loss & tie-safe pairwise ranking loss (Eq.~\ref{eq:pairwise-rank}) \\
Generator layer & final layer, cached in \texttt{float16} \\
Seeds & $\{42,123,456,789,1024\}$ for 5-seed runs; $\{42,123,456\}$ for 3-seed ablations \\
Hardware & L40S and H100 GPUs \\
\bottomrule
\end{tabular}
\end{table*}

\paragraph{Baseline configurations.}
We compare HSRM with text-based learned verifiers and non-learned generator-side
scorers. For EORM, we reproduce a Transformer energy model
over chain-of-thought text trained with a Bradley--Terry pairwise loss and a
cosine-warmup schedule. We evaluate a small EORM variant with $2.6$M parameters
($d{=}256$, $L{=}1$, $h{=}2$) and the primary EORM baseline with $53.4$M
parameters ($d{=}768$, $L{=}2$, $h{=}4$). We also evaluate
Qwen2.5-Math-PRM-7B as an off-the-shelf PRM baseline;
this model is not retrained on our sampled candidates. For Qwen2.5-Math-PRM-7B, the main results use the last-step PRM score as the
candidate-level score, matching the outcome-level target used to train HSRM.

\paragraph{Non-learned scorer definitions.}
For each candidate, let \(\ell_t = \log p_{\theta}(y_t \mid y_{<t})\) be the
token log-probability, \(H_t\) the token entropy, \(T\) the response length in
tokens, and \(S\) the number of reasoning-step delimiters. We define the
non-learned candidate scores as follows.

\begin{equation}
s_{\mathrm{cum\text{-}lp}}
=
\sum_{t=1}^{T} \ell_t .
\end{equation}

\begin{equation}
s_{\mathrm{mean\text{-}lp}}
=
\frac{1}{T}
\sum_{t=1}^{T} \ell_t .
\end{equation}

\begin{equation}
s_{\mathrm{neg\text{-}ent}}
=
-\frac{1}{T}
\sum_{t=1}^{T} H_t .
\end{equation}

\begin{equation}
s_{\mathrm{neg\text{-}varent}}
=
-\operatorname{Var}_{t}(H_t) .
\end{equation}

In Table~\ref{tab:app_cheap}, we report the best scorer separately for each
cell. This is an optimistic envelope for cheap heuristics; a single fixed scorer
would generally perform worse.

\section{Extended Results}
\label{app:extres}

\paragraph{$N$-scaling.}
Table~\ref{tab:app_nscale} varies the evaluation pool size
$N\in\{8,16,32,64\}$ while keeping the verifier fixed. Increasing $N$ gives only
small additional gains after $N{=}16$, suggesting that the sampled pools are
already close to their oracle ceiling in many cells. The MATH-500 rows use an
earlier split and are included only to show the trend.

\begin{table}[h]
\caption{\textbf{$N$-scaling of HSRM verifier-best accuracy} (\%).
$\Delta$ is the change from $N{=}8$ to $N{=}64$.}
\label{tab:app_nscale}
\centering
\footnotesize
\setlength{\tabcolsep}{3.5pt}
\begin{tabular}{lcccccr}
\toprule
Data & Gen & 8 & 16 & 32 & 64 & $\Delta$ \\
\midrule
\multirow{4}{*}{GSM8K}
 & 1.7B & 84.8 & 84.3 & 84.4 & 85.6 & $+0.8$ \\
 & 4B   & 91.5 & 92.6 & 93.1 & 92.6 & $+1.1$ \\
 & 8B   & 93.8 & 94.4 & 94.7 & 95.0 & $+1.2$ \\
 & 14B  & 95.6 & 95.5 & 95.9 & 95.8 & $+0.2$ \\
\midrule
\multirow{4}{*}{MATH-500}
 & 1.7B & 75.3 & 76.1 & 75.0 & 76.5 & $+1.2$ \\
 & 4B   & 86.0 & 86.3 & 86.2 & 86.8 & $+0.8$ \\
 & 8B   & 85.3 & 85.8 & 85.3 & 86.0 & $+0.7$ \\
 & 14B  & 87.9 & 87.8 & 87.6 & 87.9 & $\phantom{+}0.0$ \\
\bottomrule
\end{tabular}
\end{table}

\begin{table*}[t]
\caption{\textbf{Best cheap scorer vs. HSRM} (best-of-8 accuracy, \%).
\(\Delta\) is HSRM minus the best cheap scorer. The cheap scorer is selected
separately for each dataset--generator cell, so this table is an optimistic
comparison for non-learned heuristics.}
\label{tab:app_cheap}
\centering
\small
\begin{tabular}{llccr}
\toprule
Dataset & Gen & Best cheap scorer & Cheap / HSRM & \(\Delta\) \\
\midrule
GSM8K & 1.7B & neg\_varentropy & 83.2 / 85.6 & \(+2.4\) \\
      & 4B   & shortest & 92.2 / 92.6 & \(+0.4\) \\
      & 8B   & cumulative\_logprob & 94.3 / 95.0 & \(+0.7\) \\
      & 14B  & neg\_varentropy & 95.4 / 95.8 & \(+0.4\) \\
\midrule
MATH-500 & 1.7B & cumulative\_logprob & 76.4 / 77.3 & \(+0.9\) \\
      & 4B   & neg\_mean\_entropy & 86.2 / 85.8 & \(-0.4\) \\
      & 8B   & neg\_varentropy & 86.8 / 87.0 & \(+0.2\) \\
      & 14B  & shortest & 87.6 / 87.8 & \(+0.2\) \\
\midrule
AIME & 1.7B & neg\_mean\_entropy & 7.4 / 9.1 & \(+1.7\) \\
      & 4B   & shortest & 18.6 / 18.6 & \(\phantom{+}0.0\) \\
      & 8B   & cumulative\_logprob & 19.3 / 17.7 & \(-1.6\) \\
      & 14B  & cumulative\_logprob & 21.7 / 21.7 & \(\phantom{+}0.0\) \\
\midrule
OlympiadBench & 1.7B & cumulative\_logprob & 40.0 / 41.6 & \(+1.6\) \\
      & 4B   & cumulative\_logprob & 49.2 / 50.1 & \(+0.9\) \\
      & 8B   & cumulative\_logprob & 51.4 / 53.0 & \(+1.6\) \\
      & 14B  & fewest\_steps & 52.5 / 53.1 & \(+0.6\) \\
\bottomrule
\end{tabular}
\end{table*}

\begin{table*}[tb!] \centering \small \setlength{\tabcolsep}{4.8pt} 
\begin{tabular}{llcc} \toprule Dataset & Loss & Bo8 Acc. & Within-AUROC \\ \midrule \multirow{3}{*}{GSM8K} & Bradley--Terry (Eq.~\eqref{eq:pairwise-rank}) & \textbf{85.93 $\pm$ 0.56} & \textbf{0.718 $\pm$ 0.007} \\ & ListMLE & 84.59 $\pm$ 1.41 & 0.694 $\pm$ 0.027 \\ & BCE & 83.91 $\pm$ 0.45 & 0.687 $\pm$ 0.006 \\ \midrule \multirow{3}{*}{MATH-500} & Bradley--Terry (Eq.~\eqref{eq:pairwise-rank}) & \textbf{66.91 $\pm$ 0.89} & \textbf{0.620 $\pm$ 0.014} \\ & ListMLE & 66.69 $\pm$ 1.40 & 0.614 $\pm$ 0.016 \\ & BCE & 66.57 $\pm$ 2.15 & 0.587 $\pm$ 0.017 \\ \bottomrule \end{tabular} \caption{ Training-objective ablation on Qwen3-1.7B candidate pools. All runs use the same HSRM architecture; only the training objective differs. The proposed tie-safe Bradley--Terry objective achieves the best best-of-8 selection accuracy and within-problem AUROC on both datasets. } 
\label{tab:loss_ablation} 
\end{table*} 

\paragraph{Cheap scorers.}
Table~\ref{tab:app_cheap} compares HSRM against an optimistic envelope of
non-learned heuristics, where the best cheap scorer is selected separately for
each dataset--generator cell. This is a deliberately strong baseline: in a real
deployment, one would usually need to choose a single heuristic in advance,
whereas this table gives the heuristic family oracle access to the best choice
for each setting. Even under this favorable comparison, HSRM outperforms the
best cheap scorer in 12 of 16 settings, ties in 2, and underperforms in only 2.
The gains are most consistent on GSM8K and OlympiadBench, where HSRM improves
over the best heuristic at every generator scale. This suggests that the
hidden-state verifier is capturing information beyond surface-level confidence,
entropy, or length effects.

The pattern of selected cheap scorers is also informative. No single heuristic
dominates across datasets or generator scales: cumulative log-probability is
often strongest on OlympiadBench, entropy-based scores are competitive in some
GSM8K and MATH-500 settings, and length-based heuristics occasionally work well
on saturated or low-signal cells. This instability indicates that cheap scorers
mainly exploit dataset- and generator-specific artifacts rather than a robust
verification signal. In contrast, HSRM uses the same learned hidden-state
scoring mechanism across all settings.

The exceptions are concentrated on AIME and a few saturated MATH-500 cells. On
AIME, all methods operate in a low-accuracy regime with small absolute
differences, so simple confidence heuristics can sometimes match or exceed HSRM.
On saturated MATH-500 settings, the candidate pool often contains strong
solutions and the marginal benefit of learned reranking becomes smaller. Overall,
the comparison shows that HSRM is not merely learning a proxy for
log-probability, entropy, or response length. Rather, it provides a more robust
ranking signal, while cheap heuristics remain useful stress-test baselines in
regimes where confidence or length is already highly correlated with
correctness.

\begin{table*}[tbh]
\caption{\textbf{Encoder ablation} (accuracy $\pm$ standard deviation / AUROC
$\pm$ standard deviation, 5 seeds). DeepSet mean-pools per-step MLP features;
Transformer $d{=}128$ is a narrower Transformer encoder; Transformer $d{=}256$
is the default HSRM encoder.}
\label{tab:app_encoder}
\centering
\scriptsize
\resizebox{\textwidth}{!}{%
\begin{tabular}{llccc}
\toprule
Dataset & Gen & DeepSet ($d{=}128$, 0.40M) & Transformer ($d{=}128$, 0.67M) & Transformer ($d{=}256$, 2.12M) \\
\midrule
GSM8K & 1.7B & $85.76{\pm}0.57$ / $0.670{\pm}0.010$ & $85.42{\pm}0.58$ / $0.662{\pm}0.019$ & $\mathbf{85.64{\pm}0.18}$ / $0.670{\pm}0.011$ \\
      & 4B   & $92.31{\pm}0.20$ / $0.615{\pm}0.015$ & $92.50{\pm}0.47$ / $0.629{\pm}0.009$ & $\mathbf{92.55{\pm}0.46}$ / $0.638{\pm}0.013$ \\
      & 8B   & $94.77{\pm}0.47$ / $0.635{\pm}0.038$ & $94.65{\pm}0.34$ / $0.645{\pm}0.024$ & $\mathbf{94.99{\pm}0.22}$ / $0.682{\pm}0.029$ \\
      & 14B  & $95.85{\pm}0.22$ / $0.652{\pm}0.034$ & $95.80{\pm}0.17$ / $0.641{\pm}0.027$ & $95.82{\pm}0.25$ / $0.649{\pm}0.019$ \\
MATH-500 & 1.7B & $74.75{\pm}1.15$ / $0.591{\pm}0.015$ & $75.75{\pm}1.27$ / $0.568{\pm}0.012$ & $\mathbf{76.45{\pm}0.68}$ / $0.597{\pm}0.012$ \\
      & 4B   & $86.50{\pm}0.27$ / $0.586{\pm}0.016$ & $\mathbf{86.55{\pm}0.89}$ / $0.576{\pm}0.017$ & $85.80{\pm}0.78$ / $0.560{\pm}0.015$ \\
      & 8B   & $\mathbf{86.55{\pm}0.62}$ / $0.595{\pm}0.021$ & $86.55{\pm}0.43$ / $0.595{\pm}0.020$ & $85.95{\pm}1.22$ / $0.577{\pm}0.039$ \\
      & 14B  & $87.90{\pm}0.46$ / $0.568{\pm}0.021$ & $\mathbf{89.10{\pm}1.12}$ / $0.571{\pm}0.034$ & $87.90{\pm}0.78$ / $0.537{\pm}0.022$ \\
\bottomrule
\end{tabular}}
\end{table*}

\begin{table*}[t!]
\caption{\textbf{Verifier cost comparison.} ``Weights'' denotes approximate
\texttt{float16} weight memory. $S$ is the number of extracted reasoning-step
representations and $T$ is the number of generated text tokens.}
\label{tab:app_cost}
\centering
\small
\begin{tabular}{lccll}
\toprule
Verifier & Params & Weights (fp16) & Verifier input & Extra generator passes \\
\midrule
\textbf{HSRM (ours)} & \textbf{2.1--3.4M} & \textbf{$\sim$4--7MB} & \textbf{$S\le100$ hidden states} & \textbf{0} \\
EORM 55M            & 53.4M & $\sim$107MB & $T$ text tokens & text re-encoding \\
Qwen2.5-Math-PRM-7B & $\sim$7.6B & $\sim$15GB & $T$ text tokens & generator-scale re-encoding \\
\bottomrule
\end{tabular}
\end{table*}

\section{Extended Ablations}
\label{app:extabl}

\paragraph{Training Objective} \label{sec:ablations:loss} 
HSRM is trained for best-of-\(N\) selection, so its objective should align with the inference-time requirement of ranking candidates \emph{within the same problem}. This motivates the tie-safe Bradley--Terry objective in Eq.~\eqref{eq:pairwise-rank}, which encourages correct candidates to score higher than incorrect ones without imposing an arbitrary ordering among candidates that share the same binary label. To test whether this design choice matters in practice, we compare HSRM trained with three objectives: the proposed Bradley--Terry ranking loss, pointwise binary cross-entropy (BCE), and ListMLE. For a fair comparison, all variants use the same HSRM architecture and are trained on the same Qwen3-1.7B candidate pools; only the loss function is changed. Table~\ref{tab:loss_ablation} reports results on GSM8K and MATH-500.

The results favor the pairwise ranking objective. On GSM8K, Bradley--Terry reaches \(85.93\%\) best-of-8 accuracy and \(0.718\) within-problem AUROC, outperforming both ListMLE and BCE. The same pattern holds on MATH-500, where Bradley--Terry again gives the strongest ranking performance. The gains over BCE indicate that treating each candidate independently as a pointwise correctness-classification problem is less well matched to the best-of-\(N\) selection setting. The comparison with ListMLE is also informative. While ListMLE is a ranking objective, it requires a complete ordering of the candidate list. In our setting, however, multiple candidates for the same problem can be correct, especially for stronger generators. Imposing a full ordering therefore introduces unnecessary ranking pressure among candidates with the same label. The stronger performance of Bradley--Terry suggests that this tie-safe formulation is better aligned with the structure of best-of-\(N\) reasoning, where the key requirement is to rank correct candidates above incorrect ones within each problem rather than to totally order the entire candidate pool.

\paragraph{Encoder architecture.}
Table~\ref{tab:app_encoder} compares three encoder bodies while keeping the
input signal, loss, learning rate, number of training steps, batch size, and
validation split fixed. Overall, the choice of encoder has a relatively small
effect compared with the choice of input representation. DeepSet, the narrower
Transformer, and the default \(d{=}256\) Transformer all achieve comparable
accuracy, and no architecture dominates across every dataset--generator pair.
This suggests that much of the verification signal is already present in the
step-boundary hidden states, rather than being created by a high-capacity
encoder.

The differences are nevertheless informative. On GSM8K, the Transformer
encoders generally improve within-problem AUROC as generator scale increases,
with the \(d{=}256\) Transformer giving the strongest ranking quality for the
4B and 8B generators. On MATH-500, however, the pattern is less consistent:
DeepSet or the narrower Transformer can match or exceed the wider Transformer in
several cells. This suggests that increasing encoder capacity is not always
beneficial, especially when the verifier supervision is noisier or the candidate
ranking problem is more heterogeneous. We therefore use the \(d{=}256\)
Transformer as the default HSRM encoder because it is consistently competitive,
matches the architecture used in the main experiments, and provides a strong
accuracy--AUROC trade-off, while the ablation shows that HSRM's gains are not
simply due to using a larger encoder.

\section{Verifier Cost and Efficiency}
\label{app:cost}

HSRM's efficiency advantage is structural: it reads hidden states that are
already produced during generation. Therefore, after the candidates have been
sampled, HSRM requires no additional generator forward passes. In contrast,
text-based verifiers must re-encode each generated solution as text with a
separate network. Table~\ref{tab:app_cost} summarizes the resulting parameter,
memory, and input-length differences.

For HSRM, the verifier processes at most $S\le100$ step vectors with a small
2-layer Transformer encoder. Its dominant operation is the input projection,
which costs $O(S d_{\mathrm{model}} d_{\mathrm{gen}})$, followed by a small
Transformer cost in $d_{\mathrm{model}}{=}256$. Text-based verifiers instead
process the full response length $T$ and must run a separate encoder over emitted
text. For Qwen2.5-Math-PRM-7B, this means re-reading each candidate with a
billion-parameter model, making verification comparable to an additional
large-model pass over the full solution.

\end{document}